\documentclass[a4paper]{article}

\usepackage{INTERSPEECH2022}
\usepackage{url}
\usepackage{pifont}
\usepackage{subfigure}		
\usepackage{graphicx}

\usepackage{multirow}
\usepackage{float}
\usepackage{amsfonts}
\usepackage{cite}
\usepackage{cite}
\usepackage{amssymb}
\usepackage{amssymb}
\usepackage{pifont}
\usepackage{gensymb}

\usepackage{xcolor}
\usepackage{xpatch}
\title{Triple X: A LLM-Based Multilingual Speech Recognition System for the INTERSPEECH2025 MLC-SLM Challenge}
\name{Miaomiao Gao$^1$$^3$$^{*}$, Xiaoxiao Xiang$^2$$^{*}$, Yiwen Guo$^4$$^\dagger$}
\address{
  $^1$Aerospace Information Research Institute, Chinese Academy of Sciences\\
  $^2$LIGHTSPEED\\
  $^3$University of Chinese Academy of Sciences\\
  $^4$Independent Researcher\
  }
\email{xiangxiaoxiao18@mails.ucas.ac.cn, gaomiaomiao20@mails.ucas.ac.cn
}

\begin{document}

\maketitle
\renewcommand{\thefootnote}{} 
\footnotetext{%
	\begin{tabular}[t]{@{}l@{}}
		$^{*}$ Equal contribution. \\
		$^{\dagger}$ Corresponding authors.
	\end{tabular}
}
\begin{abstract}
  This paper describes our Triple X speech recognition system submitted to Task 1 of the Multi-Lingual Conversational Speech Language Modeling (MLC-SLM) Challenge. Our work focuses on optimizing speech recognition accuracy in multilingual conversational scenarios through an innovative encoder-adapter-LLM architecture. This framework harnesses the powerful reasoning capabilities of text-based large language models while incorporating domain-specific adaptations. To further enhance multilingual recognition performance, we adopted a meticulously designed multi-stage training strategy leveraging extensive multilingual audio datasets. Experimental results demonstrate that our approach achieves competitive Word Error Rate (WER) performance on both dev and test sets, obtaining second place in the challenge ranking.
  
\end{abstract}

\section{Introduction}
Speech recognition is the task of transcribing speech into text. It plays a critically important role in a wide range of applications, including human-computer interaction, voice assistants, real-time transcription, and content creation. High-accuracy speech recognition systems can significantly enhance user experience and accessibility, especially in multilingual and conversational contexts.

Classic end-to-end automatic speech recognition (ASR) frameworks have demonstrated great success in recent years. Representative methods include Paraformer~\cite{gao2022paraformer}, OWSM v3.1~\cite{peng2024owsm}, FireRedASR-AED~\cite{xu2025fireredasr}, and Whisper~\cite{radford2023robust}. These models typically follow an encoder-decoder paradigm and can be categorized into several mainstream modeling approaches: connectionist temporal classification (CTC)~\cite{graves2006connectionist}, recurrent neural network transducer (RNN-T)~\cite{graves2013speech}, Recurrent Neural Aligner (RNA)~\cite{sak2017recurrent}, and encoder-decoder approaches~\cite{chan2016listen}. All of these methods aim to learn the complex mapping between sequences of acoustic features and sequences of textual tokens by leveraging large-scale paired speech and text datasets.

Recently, text-based large language models (LLMs) have demonstrated outstanding performance across a wide range of downstream tasks, including machine translation, question answering, and long-form text generation. Models such as DeepSeek~\cite{liu2024deepseek}, GPT~\cite{achiam2023gpt}, Qwen~\cite{bai2023qwen}, and LLaMA~\cite{touvron2023llama} have emerged as foundational models for natural language understanding and generation due to their ability to capture rich linguistic and contextual knowledge from massive text corpora.
Inspired by the success of pre-trained LLMs in textual domains, recent studies have explored integrating their reasoning and generation capabilities into ASR pipelines. Notable examples include Qwen-Audio~\cite{chu2023qwen} and FireRedASR-LLM~\cite{xu2025fireredasr}, which improve performance in semantically complex or noisy conditions.
However, despite these advances, existing LLM-augmented ASR systems have not sufficiently addressed the challenges posed by real-world multilingual conversational scenarios, which involve code-switching, speaker diversity, and informal speech patterns. This highlights the need for more robust architectures and training strategies specifically tailored to multilingual, conversational ASR.

Task 1 of the MLC-SLM Competition aims to develop an LLM-based ASR system that improves speech recognition accuracy in multilingual conversational scenarios. To this end, we adopt an Encoder–Adapter–LLM architecture that leverages the capabilities of large language models (LLMs). The Encoder extracts rich acoustic and semantic representations from speech, while the adapter bridges the encoder output to the semantic space of the LLM. The LLM then generates transcriptions by interpreting both the audio-derived features and a given task instruction. By utilizing LLMs, Triple X taps into their advanced text processing capabilities and reasoning potential, enabling more accurate speech-to-text conversion and better adaptation to diverse linguistic patterns and contexts.

Using this approach, we achieve word error rates of 9.73\% and 9.67\% on the validation and test sets, respectively, securing second place on the official leaderboard. These results demonstrate the effectiveness of our architecture in improving ASR performance in multilingual settings, showing competitive results compared to other state-of-the-art models.

\section{Approach}
In this section, we first introduce the network architecture, followed by a description of the dataset used in our experiments. Finally, we detail the experimental setup, including the training strategies, input features, and loss functions.

\subsection{Network Architecture}

\begin{figure*}
	\centering
	\includegraphics[width=6.0in]{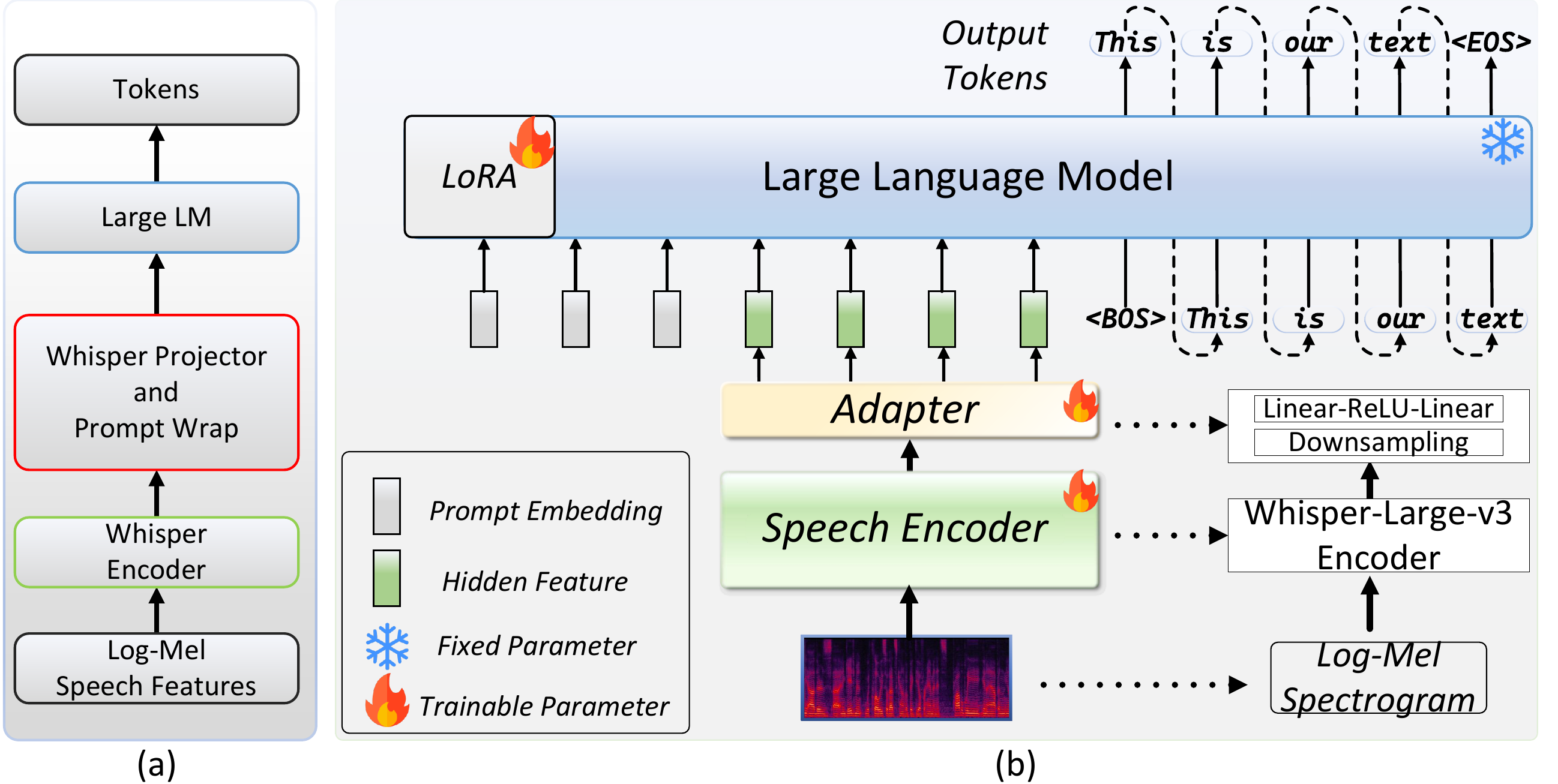}
	\caption{The schematic diagram of the Triple X architecture, which includes an encoder, an adapter and an text-based LLM.}
	\label{archi}
\end{figure*}

Our proposed Triple X system adopts the widely used Encoder-Adapter-LLM architecture, as illustrated in Figure 1. Specifically, we employ the Whisper-large-v3 encoder to extract rich acoustic and semantic features from the input speech. This encoder follows a standard Transformer architecture. However, the output sequence of the encoder is longer than that of text, which can negatively impact the processing efficiency of the LLM.
To reduce the sequence length and align the output dimension of the audio encoder with the input embedding dimension of the pre-trained text-based LLM, our adapter first applies a downsampling module to reduce the sequence length, followed by a Linear–ReLU–Linear transformation to map the output semantic information from of the encoder into the semantic space of the LLM.
Notably, we use the simplest frame splicing for the downsampling module, as we have found that different downsampling methods yield similar results.
For the LLM component in Triple X, we initialize it with pretrained weights from Qwen-3B, a high-quality open-source LLM developed by Alibaba. As shown in Figure.~\ref{archi}, the input to the LLM includes the output features of the encoder and the user prompt.

\subsection{Datasets}
\begin{figure}
	\centering
	\includegraphics[width=3.0in]{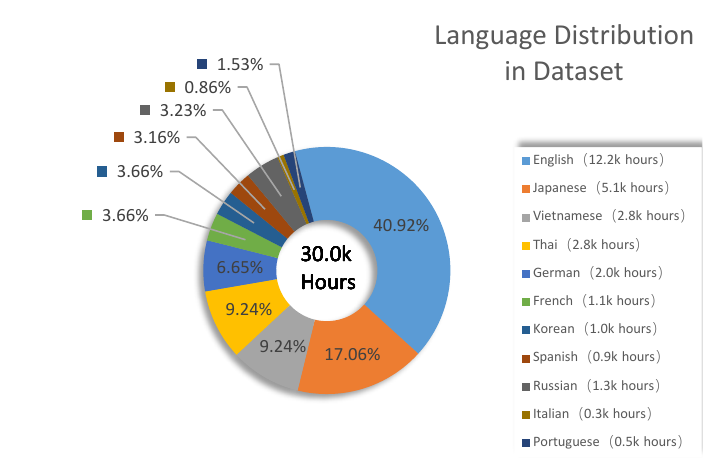}
	\caption{Illustration of the dataset distribution across different languages and their respective data volumes.}
	\label{dataset}
\end{figure}

In our experiments, we use two types of training sets. The first training set consists of approximately 1,500 hours of multilingual conversational speech data provided by the competition organizers. It covers around 11 languages, including English, French, German, Italian, Portuguese, Spanish, Japanese, Korean, Russian, Thai, and Vietnamese.
The English portion includes about 500 hours of recordings from diverse regions, including British, American, Australian, Indian, and Philippine English, while each of the other languages contributes approximately 100 hours. We apply oracle segmentation and use speaker labels for each conversation to segment long utterances into shorter ones. 
The second training set is constructed from publicly available datasets, including GigaSpeech2~\cite{ yang2024gigaspeech }, KsponSpeech~\cite{ bang2020ksponspeech }, Reazonspeech~\cite{ fujimoto2016reazonspeech }, and Multilingual LibriSpeech~\cite{ pratap2020mls }. We have selected 30,000 hours of audio data from these datasets, and statistical information regarding the amount and language of data are shown in Figure.~\ref{dataset}.

To evaluate the model, we use the development and evaluation sets provided by the competition organizers, with the evaluation set containing 4 hours of recordings for each language.

\subsection{Experiment Setup}

We adopt a carefully designed three-stage training strategy to improve multilingual speech recognition accuracy. 
First, we fine-tune Whisper-large-v3 and use the resulting encoder weights to initialize the encoder of Triple X.
This enhances the encoder’s speech feature representation capabilities and facilitates faster convergence in subsequent training stages.
Next, we freeze the encoder and train the adapter parameters to align the semantic information embedded in the encoded representations with the semantic space of the LLM.
Finally, we apply trainable Low-Rank Adaptation (LoRA) to fine-tune the LLM while keeping its core parameters fixed. This approach strikes a balance between adaptability and preservation of pre-trained knowledge.
For input speech, similar to conventional end-to-end ASR systems, we apply SpecAug~\cite{park2019specaugment} and speed perturbation~\cite{ko2015audio} for data augmentation.
We extract 128-dimensional log-Mel spectrograms as input feature of the encoder with a window of 25ms, a hop length of 10ms, without applying global mean and variance normalization. 
During training, cross-entropy loss is used and computed only at positions corresponding to the text transcription.

\begin{table}
	\centering
	\caption{Detailed WER (\%) results of different LLM backbone on the official evaluation set for various beam sizes. The \textit{\textbf{BOLD}} values show the best results.}
	\renewcommand{\arraystretch}{1.2}	
	\setlength{\tabcolsep}{0.8mm}{
		\begin{tabular}{c|c|c}
			\hline 
			System & beam size &  WER (\%) $\downarrow$\\
			\hline
			\hline		
			Qwen3-8B & 2 & 10.83 \\
			Qwen3-8B & 3 & 10.75 \\
			Qwen3-8B & 5 & 10.71 \\
			Qwen3-8B & 8 & 10.70 \\
			Qwen3-8B-Base & 2 & 10.56 \\
			Qwen3-8B-Base & 3 & 10.47 \\
			Qwen3-8B-Base & 5 & 10.43 \\
			\textbf{Qwen3-8B-Base} & 8 & \textbf{10.41} \\
			Qwen3-8B-Base & 10 & 10.42 \\
			\hline
			\hline	
	\end{tabular}}
	\label{Aba}
\end{table}

\begin{table}
	\centering
	\caption{WER(\%) results on interspeech 2025 MLC-SLM Task 1 evaluation set. The \textit{\textbf{BOLD}} values show the best results.}
	\renewcommand{\arraystretch}{1.2}	
	\setlength{\tabcolsep}{0.8mm}{
		\begin{tabular}{c|c}
			\hline 
			System &  WER (\%) $\downarrow$\\
			\hline
			\hline		
			MLC-SLM Baseline & 20.17 \\
			\textbf{Triple X} & \textbf{9.67} \\
			\hline
			\hline	
	\end{tabular}}
	\label{Aba2}
\end{table}

\section{Evaluation}

We initially optimized the model using the training set provided by the competition organizers to facilitate rapid model selection and performance validation. Table.\ref{Aba} presents the results of Qwen3-8B and Qwen3-8B-Base on the evaluation set, revealing several key insights.
First, Qwen3-8B-Base consistently achieved higher speech recognition accuracy than Qwen3-8B, as evidenced by lower WER scores across various beam settings. This suggests that the base version may serve as a more effective backbone for speech recognition tasks.
Second, increasing the beam size initially improved recognition accuracy but later led to a decline, with the best performance (lowest WER) observed at a beam size of 8. Therefore, to balance computational efficiency and recognition accuracy, we adopt a beam size of 8 as the optimal setting in subsequent experiments.

Even without incorporating additional data, the aforementioned models already achieved impressive performance, surpassing 80\% of the participants. To further enhance our results, we collected a substantial amount of publicly available datasets to better map the semantic information from the encoded representation into the semantic space of the LLM. The distribution of these datasets is illustrated in Figure~\ref{dataset}.
After pre-training, we fine-tuned both the adapter and LoRA modules using the official training set with a reduced learning rate. As shown in Table~\ref{Aba2}, the proposed method achieved a WER of 9.67\% on the official evaluation set, corresponding to a recognition accuracy of 90.33\%. This represents a 13.15\% improvement over the baseline accuracy of 79.83\%. Overall, our model achieved WERs of 9.73\% and 9.67\% on the validation and evaluation sets, respectively, securing second place on the competition leaderboard.

\section{Conclusions}

We have developed a multilingual speech recognition system named Triple X, which leverages a LLM. By employing a multi-stage training strategy, our system achieved a WER of 9.67 on the MLC-SLM evaluation set, securing second place on the leaderboard.
For future endeavors, we plan to collect more extensive multilingual conversational datasets to further enhance recognition accuracy in multilingual dialogue settings. Additionally, we aim to extend our current ASR model to support both speech recognition and response generation within a unified framework.

\bibliographystyle{IEEEtran}

\bibliography{template}


\end{document}